%% file: Eusipco_2021.tex
\documentclass[conference]{IEEEtran}
\IEEEoverridecommandlockouts
\usepackage{cite}
\usepackage{amsmath,amssymb,amsfonts,amsthm}
\usepackage{graphicx}
\usepackage{textcomp}
\usepackage{algorithm}
\usepackage{graphicx}
\usepackage[noend]{algpseudocode}
\usepackage{url}
\usepackage[usenames]{color}
\usepackage{amsfonts}
\usepackage{fancyhdr}
\usepackage{listings}
\usepackage{arydshln}
\usepackage{psfrag,caption,subcaption}
\usepackage{arydshln}

\include{macros}

\begin{document}

\title{Learning Multi-layer Graphs and a Common Representation for Clustering\\
%{\footnotesize \textsuperscript{*}Note: Sub-titles are not captured in Xplore and
%should not be used}
%\thanks{Identify applicable funding agency here. If none, delete this.}
}

\author{\IEEEauthorblockN{Sravanthi Gurugubelli and Sundeep Prabhakar Chepuri}
\IEEEauthorblockA{\textit{Indian Institute of Science, Bangalore, India} \\
sravanthig@iisc.ac.in, spchepuri@iisc.ac.in }
% \and
% \IEEEauthorblockN{Sundeep Prabhakar Chepuri}
% \IEEEauthorblockA{\textit{Department of Electrical Communication Engineering} \\
% \textit{Indian Institute of Science}\\ Bengaluru, India \\
% spchepuri@iisc.ac.in}
%\and
% \IEEEauthorblockN{3\textsuperscript{rd} Given Name Surname}
% \IEEEauthorblockA{\textit{dept. name of organization (of Aff.)} \\
% \textit{name of organization (of Aff.)}\\
% City, Country \\
% email address or ORCID}
}
%\IEEEaftertitletext{\vspace{-1\baselineskip}}
\maketitle
\begin{abstract}
In this paper, we focus on graph learning from multi-view data of shared entities for spectral clustering. We can explain interactions between the entities in multi-view data using a multi-layer graph with a common vertex set, which represents the shared entities. The edges of different layers capture the relationships of the entities. Assuming a smoothness data model, we jointly estimate the graph Laplacian matrices of the individual graph layers and low-dimensional embedding of the common vertex set. We constrain the rank of the graph Laplacian matrices to obtain multi-component graph layers for clustering. The low-dimensional node embeddings, common to all the views, assimilate the complementary information present in the views. We propose an efficient solver based on alternating minimization to solve the proposed multi-layer multi-component graph learning problem. Numerical experiments on synthetic and real datasets demonstrate that the proposed algorithm outperforms state-of-the-art multi-view clustering techniques.
\end{abstract}

\vspace*{-1mm}
\section{Introduction}
We often observe complementary information about a common source from multiple modalities or through different feature subsets in data analysis. Various aspects of interactions underlying multi-view datasets can be represented using multi-layer graphs, in which each graph layer represents a different view. All the graph layers in a multi-layer graph share the same set of nodes representing the shared entities. However,  the edges on distinct layers represent the interactions between the entities in that view~\cite{yang2018multi}. For example, in a social network graph, we can think of people's interests like favorite sport, affiliation, and hobbies as multiple views of people in the network~\cite{greene2013producing}. Each of these views can be represented by a graph giving rise to a multi-layer graph with each layer consisting of the same set of nodes representing people. Also, one can have data from multiple modalities. For example, in brain imaging and analysis, data from different modalities, like fMRI, sMRI and DTI~\cite{Calhoun2016mrifusion}, may be considered as signals on a multi-layer graph. The edges in different graph layers represent interactions between the same set of brain regions observed in different modalities~\cite{battiston2017multilayer,vaiana2018multilayer}.  

The underlying graph structure is leveraged to solve several signal processing and machine learning tasks like denoising, spectral clustering, and dimensionality reduction. However, the underlying graph may not always be readily available. Learning graphs from data is an ill-posed problem. Although nearest neighbor graphs, correlation graphs, or Gaussian similarity kernels learnt from data are simple and commonly used choices, they are sensitive to noise or missing samples. Hence, several graph learning solutions that carefully model the data (e.g., using a smoothness or probabilistic graphical model) and incorporate prior information (such as sparsity, product, or multi-component structure) about the resulting topology have been proposed~\cite{Dong1,dong19LearnGraphData,mateos19connectingdots, coutino2018sparsest, chepuri2018distributed, kadambari2020learning,SPCGraphLearning}. Although the existing graph learning methods can be directly used to learn the individual graph layers of a multi-layer graph, the information contained in all the views cannot be captured by any of the individual graph layers. Therefore, we propose multi-layer graph learning from multi-view data in this paper.

Low-dimensional embeddings of the nodes of a graph that encode the structural information about the graph are useful in graph-based spectral clustering and other graph analysis tasks~\cite{hamilton2017representation, SpectralClustering}. A common technique to compute these low-dimensional node embeddings is to perform a partial eigenvalue decomposition of the graph underlying the data. When dealing with multi-layer graphs, it is necessary to find low-dimensional embeddings of the nodes common to all the views that best capture the complementary information available in different views. 
Assuming that the individual graph layers of a multi-layer graph are available, there are subspace learning methods that learn a common subspace by merging the multiple subspaces computed from the individual graph layers~\cite{Dong2,Dong3}. In the context of multi-view canonical correlation analysis (MCCA),~\cite{GMCCA} learns a common subspace that is smooth on a known graph from multi-view data. 

Instead, in this paper, we jointly learn a multi-layer graph and common low-dimensional node embeddings from multi-view data for multi-view clustering. Specifically, given $M$ views of data, we estimate a graph with $M$ layers having $K$ components. The respective views of data are smooth on the graph topology of individual layers that we learn. The graph is restricted to a $K$-component graph to cluster data in $K$ groups by imposing rank constraints on the graph Laplacian matrices of the individual graph layers. We propose an efficient solver based on alternating minimization, each subproblem of which is solved optimally. We evaluate the performance of the proposed method for clustering on synthetic and real-world datasets and compare them with state-of-the-art techniques for multi-view clustering~\cite{Dong2,Dong3,GMCCA}. The results show that our algorithm outperforms single-view clustering that ignores information from different views and state-of-the-art multi-view clustering methods.

\section{Multi-layer graphs} \label{sec:multilayergraph}

Consider a $M$-layer weighted and undirected graph $\cG$ with individual graph layers
$\cG_m= \{\cV,\mW_m\}$, $m = 1,2, \ldots, M$, where $\cV = \{v_1, \cdots v_N\}$ denotes the common vertex (or node) set with $N = |\cV|$ nodes and 
$\mW_{m} \in \mathbb{R}^{N \times N}$ denotes the weighted adjacency matrix of the $m$th graph layer $\cG_{m}$. The $(i,j)$th element of $\mW_{m}$ contains a positive edge weight if two nodes $v_i$ and $v_j$ are connected in $\cG_m$ and is zero otherwise. The adjacency matrix $\mW_{m}$ is symmetric as $\cG_{m} $ is undirected. The degree of the nodes in ${\cG}_{m}$ is defined as ${\mW_m}{\bf 1}$. The combinatorial graph Laplacian matrix $\mL_{m} \in \mathbb{R}^{N \times N}$ for $\cG_{m}$ is defined as $\mL_m = \diag[{\mW_m}{\bf 1}] - \mW_{m}$. By construction, $\mL_m$ is a symmetric, positive semidefinite matrix, and has a zero row sum. The space of all the valid combinatorial graph Laplacian matrices of size $N$ is given by
\begin{equation}
\label{eq:Lapspace}
\mathcal{L}=\{\mathbf{L} \in \mathbb{S}_+^{N} \mid \mathbf{L} \mathbf{1}=\mathbf{0}; L_{ij}=L_{ji} \leq 0, \,\, \forall i \neq j \},
\end{equation}
in which the constraint $L_{ii} \geq 0,\, i=1, 2, \cdots, M$ is implicit. Here, $\mathbb{S}_+^{N}$ is the set of positive semidefinite matrices of size $N \times N$.

The number of connected components in a graph is given by the multiplicity of the zero eigenvalue of its graph Laplacian matrix~\cite{SpectralClustering}. Thus the rank of the graph Laplacian matrix of a $K$-component graph with $N$ nodes is $N-K$. The eigenvectors of the graph Laplacian matrix corresponding to the $K$ zero eigenvalues preserve the nodal connectivity information and hence are used as low-dimensional node embeddings.  

 A signal (or dataset) indexed using the nodes of a graph is referred to as a graph signal (or data). Consider $M$ datasets  $\mX_m \in \mathbb{R}^{N \times D_m}$, $m=1,2,\ldots,M$ obtained from $M \geq 2$ views of a common source. The $n$th row of $\mX_m$ that contains $D_m$ features of the $n$th entity resides on the $n$th node of $\cG_m$. That is, we interpret the $M$-view dataset as a graph signal defined on the multi-layer graph $\cG$ with each layer representing a different view of the dataset. 
 
We use smoothness to measure how well a signal matches the underlying graph. A signal is said to be smooth over a graph if the signal values residing on adjacent nodes having large edge weights are similar. For the signal $\mX_{m} \in \mathbb{R}^{N\times D_m}$ residing on $\cG_m$ with the graph Laplacian matrix $\mL_m$, the smoothness is measured using the total variation with respect to the graph $\cG_m$, and is given by $\tr (\mX_{m}\rT\mL_{m}\mX_{m}) = \tr(\mL_m\mS_m)$. Here, $\mS_m = \mX_m\mX_m\rT$ is the sample data covariance matrix.
\vspace{-1mm}
\section{Problem statement} \label{sec:ProblemStatement}
%\vspace{-2.5mm}
In this work, we propose a {\it rank-constrained multi-layer graph learning} technique for clustering multi-view data. Specifically, we estimate the graph Laplacian matrices $\{\mL_{m}\}_{m=1}^M$ that best explain the $M$-view dataset $\{\mX_m\}_{m=1}^M$ by assuming that each view of the dataset is smooth on the graph of that layer. By constraining the rank of $\mL_m$, $m=1,2,\ldots,M$ to $R = N-K$, we get $K$-component graph layers. Thus partitioning the nodes into $K$ clusters. Since the entities in the multi-view dataset correspond to the same nodes in all the views, there exists a common low-dimensional representation of the nodes. To compute the common low-dimensional node embeddings, we propose {\it joint diagonalization} of $\{\mL_m\}_{m=1}^M$: 
\begin{align}
\mL_m = \mU \diag(\blambda_m) \mU\rT
= \left[\begin{array}{c|c}\mQ & \star \end{array}\right]\left[\begin{array}{c|c}{\bf 0} &      \\ \hline  & \star \end{array}\right]
\left[\begin{array}{c}\mQ\rT \\ \hline \star\end{array}\right], 
\label{eq:jointdiag}
\end{align}
for $ m = 1,2,\ldots, M$, 
where the columns of $\mU$ contain the joint eigenvectors (common factors) shared by all the graph layers and   
the vector $\blambda_m = [\lambda_1(\mL_m), \lambda_2(\mL_m), \cdots,\lambda_N(\mL_m)]\rT \in \mathbb{R}_+^N$ contains the eigenvalues of $\mL_m$. Here, $\lambda_i(\mL_m)$ is the $i$th eigenvalue of $\mL_m$, where we assume $0 = \lambda_1(\mL_m) \leq \cdots \leq \lambda_N(\mL_m)$. The common eigenvectors corresponding to the $K$ zero eigenvalues of $\{\mL_m\}_{m=1}^M$ are collected in $\mQ \in \mathbb{R}^{N \times K}$. 

To estimate the graph Laplacian matrices corresponding to the graphs in each layer, we propose the following {\it rank-constrained multi-layer graph learning} (\texttt{RMGL}) optimization problem:
\begin{align}
& \underset{\{\mL_m\}_{m=1}^M}{\rm{minimize}}
& & \sum_{m=1}^{M}\tr( \mL_m\mS_m)+ \alpha_m \left\|\mL_m\right\|_{F}^{2} \nonumber\\
& \text{subject to}
& & \,\,\mL_{m} \in \mathcal{L},  \quad \tr(\mL_m)=N,   \nonumber\\
& && \,\,\mathrm{rank}(\mL_{m})=R, \quad \, m=1,\cdots,M,
\label{eq:rmgl}
\end{align}
where recall that the set $\mathcal{L}$, defined in \eqref{eq:Lapspace}, is the set of all the valid combinatorial graph Laplacian matrices. The first term in the objective function promotes smoothness of $\mathbf{X}_m$ with respect to the graph corresponding to the $m$th layer. The second term in the objective function with the tuning parameter $\alpha_m > 0$ allows us to control the sparsity (i.e., the number of nonzero entries) of $\mW_m$. The trace constraint of the form $\tr(\mL_m)= 2 \| {\rm vec}(\mW_m)\|_1= N$ fixes the scale of the solution and avoids the trivial solution (more details in Section~\ref{sec:UpdateL}). The rank constraints on the Laplacian matrices promote $K$-component graph layers. 

For $\mL_m \in \mathbb{S}_+^N$, we have~\cite{tao2012topics}
\begin{equation}
\sum\limits_{i=1}^K \lambda_i(\mL_m) = \underset{\mQ \in \mathbb{R}^{N \times K}, \mQ^T\mQ = \mI_K}{\rm minimize} \quad \tr({\mQ^T\mL_m\mQ}).
\label{eq:KyFan_knorm}
\end{equation}
The optimal $\mQ$ is given in \eqref{eq:jointdiag}. Using this property, the rank constraints in \eqref{eq:rmgl} can be replaced with a sum-of-smallest-eigenvalues regularizer in the objective function as
\begin{align}
& \underset{\{\mL_m\}_{m=1}^M, \mQ}{\rm{minimize}}
& & \sum_{m=1}^{M}\tr( \mL_m\mS_m)+ \alpha_m \left\|\mL_m\right\|_{F}^{2} + \beta_m \tr({\mQ^T\mL_m\mQ}) \nonumber\\
& \text{subject to}
& & \,\,\mL_{m} \in \mathcal{L},  \quad \tr(\mL_m)=N, \quad  m=1,\cdots,M, \nonumber\\
& && \,\, \mQ^{T}\mQ = \mI_K, 
\label{eq:finalprob}
\end{align}
where for sufficiently large $\beta_m > 0$,  we achieve ${\rm rank}(\mL_m) = N-K$ for $m=1,2,\ldots,M$. The columns of the isometry $\mQ \in \mathbb{R}^{N \times K}$ form an orthonormal basis for the low-dimensional subspace common to the $M$ layers of the graph. In particular, the $n$th row of $\mQ$ corresponds to the $K$-dimensional embedding of the $n$th node. The  problem \eqref{eq:finalprob} is non-convex in the variables $\{\mL_m\}_{m=1}^M \text{and } \mQ$. In what follows, we present an efficient algorithm to solve for the unknowns $\{\mL_m\}_{m=1}^M \text{and } \mQ$ given the multi-view dataset $\{\mS_m\}_{m=1}^M$.
\vspace*{-1mm}
\section{Proposed solver} \label{sec:solver}

In this section, we solve the problem in \eqref{eq:finalprob} by alternatingly minimizing it with respect to $\{\mL_{m}\}_{m=1}^{M}$ and $\mQ$, while keeping the other variable fixed.
\vspace*{-2mm}
\subsection{Update of $\{\mL_m\}_{m=1}^M$}\label{sec:UpdateL}

Given $\mQ$, the problem \eqref{eq:finalprob} simplifies to the following convex optimization problem
\begin{align}
& \underset{\{\mathbf{L}_{m}\}_{m=1}^M}{\text{minimize}}
& & \sum_{m=1}^{M}\tr(\mathbf{L}_m \mR_m)+\alpha_{m}\left\|\mathbf{L}_m\right\|_{F}^{2}  \nonumber \\
& \text{subject to}
& & \mL_m \in \mathcal{L}, \quad \tr(\mL_{m})=N, \quad m=1,\cdots,M,
\label{eq:cvx_L}
\end{align}
where we have introduced the $N \times N$ matrix $\mR_m = \mS_m + \beta_m \mQ\mQ\rT$. It can be seen that the parameter $\beta_m$ regularizes the data with the eigenvectors of the graph Laplacian. Since $\mathbf{L}_{m}$ is a symmetric matrix, this quadratic program can be solved very efficiently by solving only for the {\it upper triangular entries} of $\mL_m$ as described next.

Let us define a duplication matrix $\mD \in \mathbb{R}^{N^2\times V}$ with $V = N(N+1)/2$ as $\mD\rT = \sum_{i \geq j} {\boldsymbol \delta}_{ij} {\rm vec}\rT({\boldsymbol \Theta}_{ij})$, 
where the vector ${\boldsymbol \delta}_{ij} \in \mathbb{R}^{V}$ has $1$ at position $(j-1)N + i - \frac{1}{2}j(j-1)$ and zero elsewhere. The matrix ${\boldsymbol \Theta}_{ij} \in \mathbb{R}^{N \times N}$ has $-1$ at the positions $(i,j)$  and  $(j,i)$, $1$ at $(i,i),$ and zero elsewhere. Let us collect the absolute values of the $V$ nonduplicated entries of $\mL_m$ in $\vl_m \in \mathbb{R}_+^V$. Then we have 
$
{\rm vec}(\mL_m) = \mD\vl_m,\quad m=1,2,\ldots,M,
$
where ${\rm vec}(\cdot)$ denotes the matrix vectorization operator.

Using this transformation, we express the two equality constraints in \eqref{eq:cvx_L}, namely, $\tr(\mL_{m}) = \mathrm{vec}(\mI)^T \mathrm{vec}(\mL_{m}) = {N}$ and $\mathbf{L} \mathbf{1} = (\mathbf{1}^T \otimes \mI) \mathrm{vec}(\mL_{m}) =\mathbf{0}$ as $\mC\vl_m = \vd$, where $\mC = [\mD\rT{\rm vec}(\mI), \mD\rT(\mathbf{1} \otimes \mI)]\rT \in \mathbb{R}^{N+1 \times V}$ and $\vd = [N, \mathbf{0}]^T  \in \mathbb{R}^{N+1}$. Here, $\otimes$ denotes the Kronecker product. Similarly, we express the first term in the objective function of \eqref{eq:cvx_L} as $\tr(\mathbf{L}_m \mR_m) = {\rm vec}\rT(\mR_m) {\rm vec}(\mathbf{L}_m) = \vr_m\rT\vl_m$. The second term in the objective function of \eqref{eq:cvx_L} simplifies to $\alpha_m\|\mL_m\|_F^2 = \alpha_m{\rm vec}(\mathbf{L}_m)\rT{\rm vec}(\mathbf{L}_m)$ $=\frac{1}{2}\vl_m\rT \diag(\vp_m) \vl_m$, where we have used the fact that $\mD\rT\mD$ is a positive definite diagonal matrix to obtain $ \diag(\vp_m) = 2\alpha_m\mD\rT\mD$. Now, we can simplify \eqref{eq:cvx_L} as
\begin{align}
& \underset{\{\vl_{m}\}_{m=1}^M}{\text{minimize}}
& & \sum_{m=1}^{M} \frac{1}{2}\vl_m\rT \diag(\vp_m) \vl_m + \vr_m\rT\vl_m  \nonumber \\
& \text{subject to}
& & \mC\vl_m = \vd , \quad \vl_m \succeq {\bf 0}, \quad m=1,\cdots,M.
\label{eq:cvx_lsym}
\end{align}
This is a special form of a convex quadratic program in which the matrix associated with the quadratic term is diagonal. It is computationally efficient and equivalent to solve for each $\vl_m$, $m=1,2,\ldots,M$ separately as this convex program is separable in these variables.

The Lagrangian function for the problem \eqref{eq:cvx_lsym} associated to the variable $\vl_m$  is given by
\begin{align*}
\cJ(\vl_m,\,\vgl_m,\,\vgm_m) &= \frac{1}{2} \vl_m\rT \diag(\vp_m) \vl_m  \nonumber\\ 
& \hskip5mm+ \vr_m\rT \vl_m + \vgm_m\rT(\vd -\mC\vl_m) - \vgl_m\rT \vl_m, \nonumber
\end{align*}
for $m=1,2,\ldots,M$, where $\vgm_m \in \mathbb{R}^{N+1}$ and $\vgl_m \in \mathbb{R}^V$ are the Lagrange multipliers corresponding to the equality and inequality constraints, respectively. The Karush-Kuhn-Tucker (KKT) conditions are given by
\begin{equation*}
\begin{aligned}
&\diag(\vp_m) \vl_m^\star + \vr_m - \mC\rT \vgm_m^\star - \vgl_m^\star = {\bf 0}, \\
&\,\,\mC \vl_m^\star = \vd, \quad \vl_m^\star \succeq {\bf 0},  \quad \vgl_m^\star \odot \vl_m^\star = {\bf 0},
\end{aligned}
\end{equation*}
where $\odot$ denotes the elementwise Hadarmard product. Eliminating the variable $\vgl_m^\star$ 
and solving for $\vl_m^\star$, we get
$
\vl_m^\star(\vgm_m^\star) =\left\{\diag^{-1}(\vp_m)[\mC\rT\vgm_m^\star - \vr_m] \right\}_+,
$
where $\{\cdot\}_+$ denotes the elementwise projection onto the nonnegative orthant.
Using $\vl_m^\star(\vgm_m^\star)$ in the second KKT condition, we can compute $\vgm_m^\star$ iteratively as
\begin{align}
& \vl_m^{(k)}  =  \left\{\diag^{-1}(\vp_m)[\mC\rT\vgm_m^{(k)} - \vr_m] \right\}_+,\\
&\vgm_m^{(k+1)} = \vgm_m^{(k)} - \rho [ \mC\vl_m^{(k)} - \vd], 
\end{align}
where $\rho > 0$ is the step size. We initialize the iterations with $\vgm_m^{(0)}$. The computational complexity of these iterations is dominated by the matrix-vector multiplication. Since $\mC$ is sparse with $N(N+1)$ non-zero entries, the above iterative procedure approximately costs order $N^2$ flops. For $M$ variables, the total computation cost is approximately order $MN^2$ flops.
\vspace*{-1mm}
\subsection{Update of $\mQ$} \label{subsec:Update for Q}

Given $\{\mL_{m}\}_{m=1}^M$, \eqref{eq:finalprob} reduces to the following eigenvalue problem
\begin{align}
& \underset{\mathbf{Q} \in \mathbb{R}^{N \times K}}{\text{minimize}} \quad
\tr(\mathbf{Q}^{T}\mL\mQ) \nonumber \\
& \text{subject to} \quad \mQ^{T}\mQ = \mI_K \quad \text{and} \quad \mL = \sum_{m=1}^{M}\beta_{m}\mL_{m}.
\label{eq:15}
\end{align}
The optimal $\mQ$ is given by the $K$ eigenvectors corresponding to the $K$ smallest eigenvalues of $\mL = \sum_{m=1}^{M}\beta_{m}\mL_{m}$. Computing this partial eigendecomposition approximately costs $KN^2$ flops~\cite{saul2000introduction}.

The complete alternating minimization procedure is summarized as Algorithm~\ref{alg:algorithm1}. It approximately costs order $(M+K)N^2$ flops per iteration. Furthermore, it can be shown that each limit of the sequence of updates of the optimization problems in (7) and (10) satisfy the KKT conditions of (5). 

%% Next section
\begin{figure*}
    \centering
    \begin{minipage}{0.4\textwidth}
        \centering
        \includegraphics[width=1\textwidth]{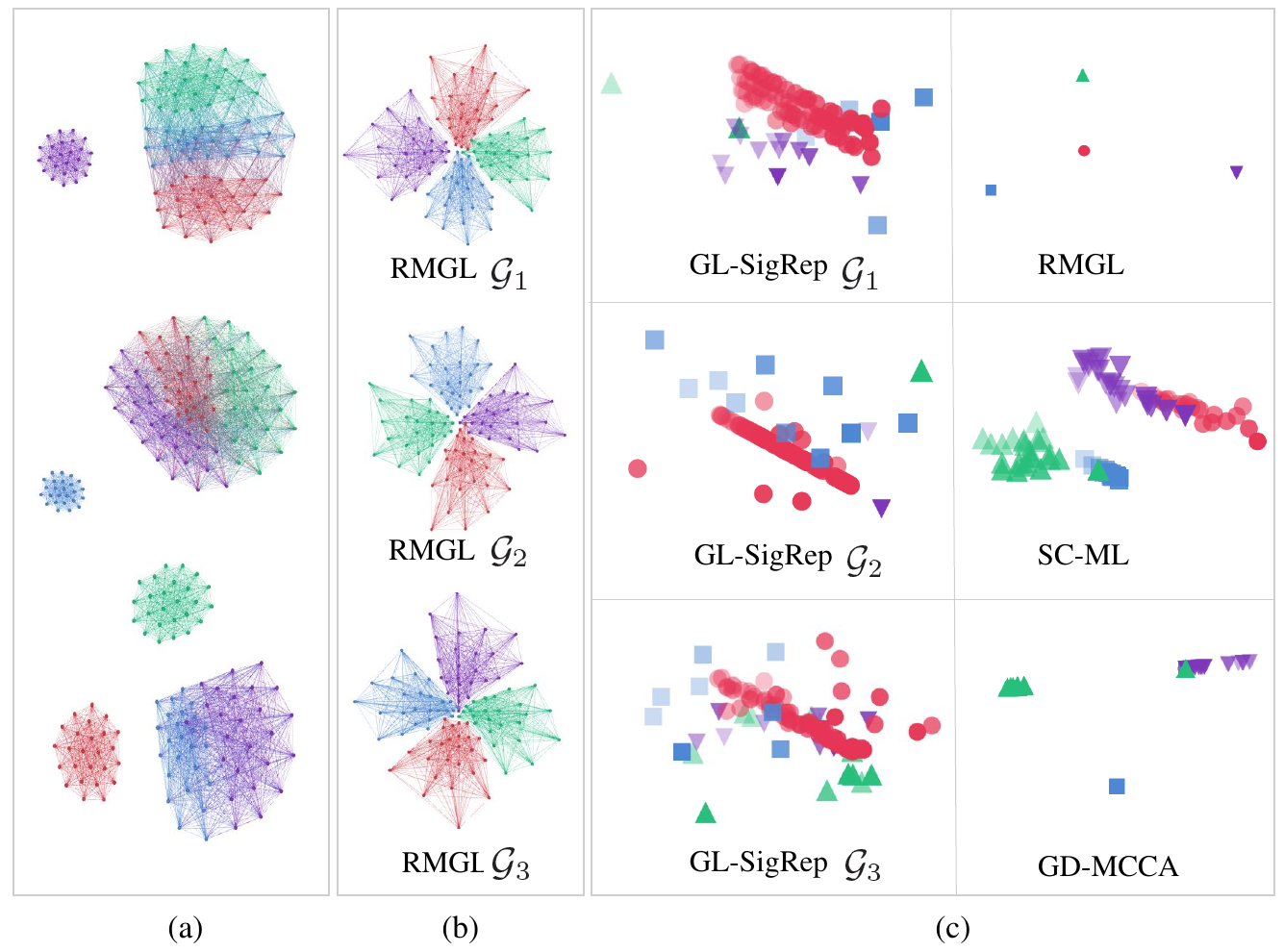}
        %\vspace{4mm}
        \caption{(a) Ground truth graphs. (b) Reconstructed graphs. (c) Node embeddings.}
        \label{fig:synthetic}
        \vspace{-5mm}
    \end{minipage}\hfill
    \begin{minipage}{0.58\textwidth}
        \centering
        \vspace*{7mm}
        \includegraphics[width=1\textwidth]{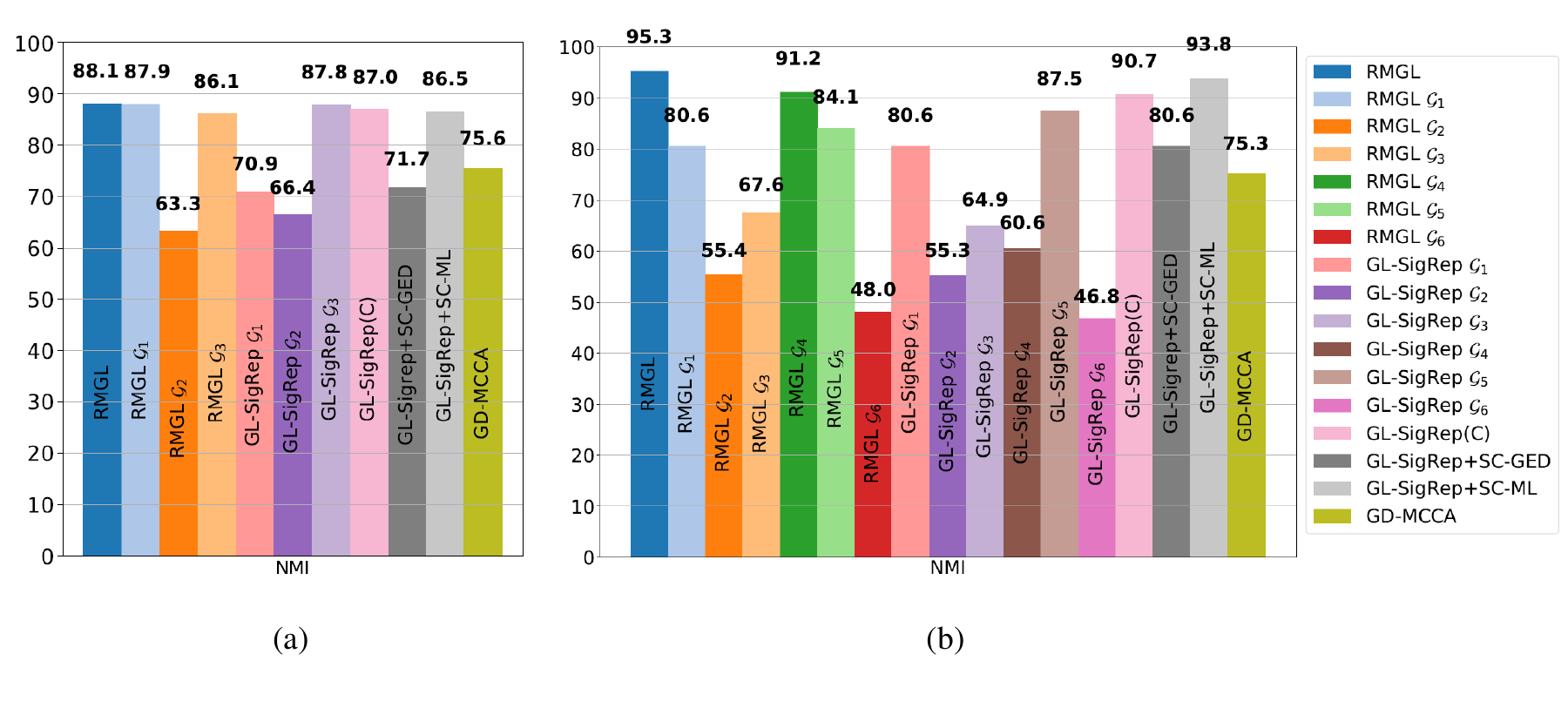}
        \caption{NMI. (a) UCI hand-written digits (b) COIL-20 datasets.}
        \label{fig:real}
    \end{minipage}
    \vspace{-1mm}
\end{figure*}

\begin{algorithm}[!b]
\caption{Rank-constrained multi-layer graph learning}\label{alg:algorithm1}
\begin{algorithmic}[1]
\Function{\texttt{RMGL}}{$\{\vp_m, \vr_m\}_{m=1}^M$, $\mC$, $\mD$, $\vd$, $K$, \texttt{Tol}, $\rho$, $\texttt{MaxIter}$}

\State Initialize $k \gets  0$ 
%\Comment{We may initialize with an all-zero or a random vector}
\While{ $k <  \texttt{MaxIter}$} 
\For{$m=1$  to $M$}
\State Initialize $\vgm_m \gets 0$ 
%\Comment{We may initialize with an all-zero or a random vector}
\While{ $\|\mC\vl_m - \vd\|_2 <  \texttt{Tol}$} 
\State $\vl_m  \gets  \left\{{\rm diag}^{-1}(\vp_m)[\mC^T\vgm_m - \vr_m] \right\}_+$
\State  $\vgm_m \gets \vgm_m - \rho [ \mC\vl_m - \vd]$
\EndWhile
\State $\mL_m \gets \texttt{mat}(\mD\vl_m)$ 

\Comment {\footnotesize$\texttt{mat}$ is the inverse vectorization operator.}

\EndFor
\State $\mL \gets \sum_{m=1}^M \beta_m \mL_m$
\State $\mQ \gets \texttt{eigs}(\mL, K)$

\Comment {\footnotesize\texttt{eigs} computes the eigenvectors associated to the $K$ smallest eigenvalues of the matrix argument.}

\State $k \gets k +1$

\EndWhile

\Return $\{\mL_m\}_{m=1}^M$ and $\mQ$

\EndFunction
\end{algorithmic}
\end{algorithm}
%\vspace*{-1mm}
\section{Numerical experiments}\label{sec:Numerical Exp}

In this section, we evaluate our framework in terms of clustering performance on a synthetically generated dataset and two real-world datasets. We use normalized mutual information (NMI) as a measure to assess the clustering performance. We compare the clustering performances of the proposed method \texttt{RMGL} with both common baseline methods and state-of-the-art multi-view spectral clustering methods. The single-view graph learning method based on signal representation (\texttt{GL-SigRep}) in~\cite{Dong1} serves as our first baseline method. We learn the graph layers independently for each of the views available using \texttt{GL-SigRep}. We also learn a graph Laplacian matrix by giving the concatenated views $\mX \in \mathbb{R}^{N\times\sum_{m=1}^M D_m}$ as input to \texttt{GL-SigRep}. We refer to the second baseline method that uses concatenated data as input as \texttt{GL-SigRep(C)}. Spectral clustering is then performed to find the clusters on the graph Laplacian matrices estimated using the above mentioned two methods. We also compare our method with state-of-the-art methods for multi-view spectral clustering, which are based on a two-step approach of graph learning followed by subspace merging. Graph-regularized dual multi-view CCA (\texttt{GD-MCCA})~\cite{GMCCA} learns a low-dimensional representation of the data common to all the views by minimizing the distance between the canonical variables and the common low-dimensional representations while leveraging a graph structure of the common source. We compute the common graph required by \texttt{GD-MCCA} as in~\cite{GMCCA}, by taking each of the affinity matrices $\mD_m^{-1}\mX_m\mX_m\rT$ as the common graph. Here, $\mD_m$ is the degree matrix of the similarity graph $\mX_m\mX_m^T$. We finally consider the common graph giving the best results for comparison. The multi-view spectral clustering methods, \texttt{SC-GED}~\cite{Dong2} and \texttt{SC-ML}~\cite{Dong3}, obtain the common node embeddings by computing a joint spectrum via a generalized eigendecomposition and merging the node embeddings from multiple graph layers, respectively. We give the independent graph layers estimated from~\cite{Dong1} as input to \texttt{SC-GED} and \texttt{SC-ML} to find the common node embeddings and perform spectral clustering.

We first curate a synthetic dataset to illustrate the merit of the proposed method through a pedagogical approach. We construct a 3-layer graph, as shown in Fig.~\ref{fig:synthetic}(a). Each of the graph layers consists of $N=100$ nodes with an equal number of samples from four classes, indicated by different colors. The graph layers are so constructed that they carry complementary information that none of the individual graphs provide. Each layer has a different class of nodes disconnected from the rest of the nodes that have inter-class connections. The denser within-class connections than the inter-class connections present in all the views convey that the nodes with different colors belong to different classes. We generate data $\mX_m \in \mathbb{R}^{100 \times 15}$, $m=1,2,3$ by taking the eigenvectors corresponding to the 15 smallest eigenvalues of the individual graph Laplacian matrices and corrupting it with zero-mean additive white Gaussian noise of variance 0.01. The estimated graph layers with $K$-components from the proposed method \texttt{RMGL} are shown in Fig.~\ref{fig:synthetic}(b). From Fig.~\ref{fig:synthetic}(c), we can see that the embeddings obtained using the three estimated graph layers using \texttt{GL-SigRep} fail to incorporate the combined information from different views. In contrast, \texttt{RMGL} assigns the same node representations to the points belonging to one class and hence are better clustered by our proposed method than the representations obtained using \texttt{SC-GED}, \texttt{SC-ML}, and \texttt{GD-MCCA}.

\texttt{RMGL}'s performance on the real datasets, UCI hand-written digits~\cite{duin1998uci} and COIL-20~\cite{nane1996columbia}, shows its merit more accurately. The UCI dataset consists of 200 instances of images of the 10 digits 0 to 9. We consider six different views of this data as in~\cite{GMCCA} with $N= 2000$, $D_1 = 216$, $D_2 =76$, $D_3 = 64$, $D_4 = 6$, $D_5 = 240$, and $D_6 = 47$. The COIL-20 dataset is an image dataset consisting of 20 classes of objects with 72 images of each object captured at different angles. We construct three views of this data related to the local binary pattern, histogram of oriented gradients, and pixel values of resized images with $N= 1440$, $D_1 = 59$, $D_2 = 36$, and $D_3 = 1024$. 

\begin{figure}
    \centering
    \includegraphics[width = 0.8\columnwidth]{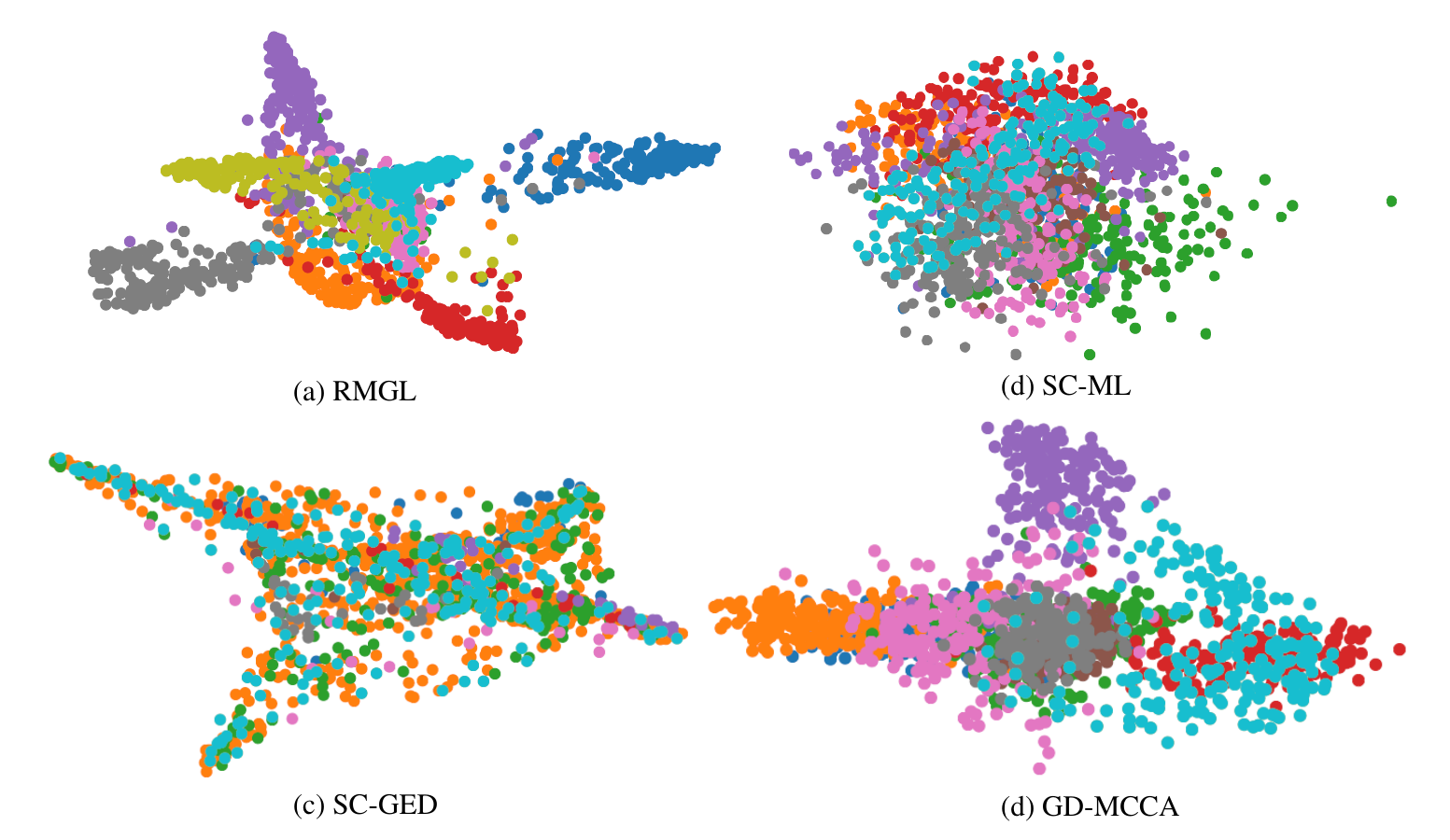}
    \vspace*{-2mm}
    \caption{Node embeddings of the UCI hand-written digits dataset.}
    \label{fig:embeddings}
    \vspace*{-6mm}
\end{figure}

With the different views of data taken as input, we estimate $\{\mL_m\}_{m=1}^M$ and $\mQ$.  The rows of $\mQ$ are then given as input to the $k$-means algorithm for clustering. We also perform spectral clustering on $\{\mL_m\}_{m=1}^M$. The accuracy of the clusters obtained using these can be treated as a measure of the correctness of the estimated graph. 
The parameters involved in the different methods considered are chosen to obtain the best possible NMI. The values of $\{\beta_m\}_{m=1}^M, $ for \texttt{RMGL}, can be chosen according to the importance of the available views by choosing a higher value of $\beta_m$ for a view $m$ that is more informative. Our model gives the best results for the following choice of parameters:  $\alpha_m = 100$, $m=1,\ldots, 6$, $\beta_1=\beta_2=\beta_3 =2$, and $ \beta_4=20 , \beta_5=7, \beta_6 = 1$ for the UCI dataset. For the COIL-20 dataset, we use $\alpha_m= 100$, $m=1,2,3$, and  $\beta_1=12, \beta_2=2, \beta_3 = 15$. The bar plot in Fig.~\ref{fig:real} shows the NMI values averaged over 20 experiments. The NMI variance using the proposed and competing methods is very small (about $10^{-6}$), and hence not shown. The NMI scores of the different methods in Fig.~\ref{fig:real} suggest that using $\mQ$ from \texttt{RMGL} as an input to the $k$-means algorithm gives the best clustering accuracy when compared to the other considered methods (including performing spectral clustering on the estimated graph layers either from \texttt{RMGL}, \texttt{GL-SigRep} or \texttt{GL-SigRep(C)}).  Fig.~\ref{fig:embeddings} shows the node embeddings corresponding to the 2000 samples from the ten classes of data in the UCI dataset set. We show the two-dimensional node representations obtained from the last two columns of the common $K$-dimensional subspaces computed using \texttt{RMGL}, \texttt{SC-GED}, \texttt{SC-ML}, \texttt{GD-MCCA}. The embeddings obtained from \texttt{RMGL} show better separability, with the node embeddings from the same cluster concentrated and those from different clusters far apart.
%% Next section
\vspace*{-1mm}
\section{Conclusions} \label{sec:conclusions}
We developed a framework for multi-view clustering, where we simultaneously learn a multi-layer graph with a common vertex set and a low-dimensional representation for the nodes common to all the views. We presented an efficient solver based on an alternating minimization procedure to solve the proposed non-convex optimization problem. We demonstrated via numerical experiments on synthetic and real datasets that the proposed method performs better than the existing multi-view clustering methods based on CCA or merging the node embeddings of the individual graph layers.

\bibliographystyle{IEEEtran}
\bibliography{references}
\end{document}

%% file: macros.tex
\def\cast{{
   \mathord{
      \hbox to 0em{
         \ooalign{
	   \smash{\hbox{$\ast$}}\crcr
	   \smash{\hskip-1pt\Large\hbox{$\circ$}} }
	 \hidewidth}
      \phantom{\bigcirc}
} }}
\def\bm#1{\mbox{\boldmath $#1$}}

\newcommand{\vgl}{\mbox{$\bm \lambda$}}

\newcommand{\vgm}{\mbox{$\bm \mu$}}

\newcommand{\rT}{^{ \raisebox{1.2pt}{$\rm \scriptstyle T$}}}

\newcommand{\bds}{\begin {itemize}}
\newcommand{\eds}{\end {itemize}}
\newcommand{\bdf}{\begin{definition}}
\newcommand{\blm}{\begin{lemma}}
\newcommand{\edf}{\end{definition}}
\newcommand{\elm}{\end{lemma}}
\newcommand{\bthm}{\begin{theorem}}
\newcommand{\ethm}{\end{theorem}}
\newcommand{\bprp}{\begin{prop}}
\newcommand{\eprp}{\end{prop}}
\newcommand{\bcl}{\begin{claim}}
\newcommand{\ecl}{\end{claim}}
\newcommand{\bcr}{\begin{coro}}
\newcommand{\ecr}{\end{coro}}
\newcommand{\bquest}{\begin{question}}
\newcommand{\equest}{\end{question}}

\newcommand{\larrow}{{\larrow}}

\newcommand{\argmin}{\ensuremath{\mathrm{arg}\min}}
\newcommand{\argmax}{\ensuremath{\mathrm{arg}\max}}

\newcommand{\cG}{{\ensuremath{\mathcal{G}}}}

\newcommand{\cJ}{{\ensuremath{\mathcal{J}}}}

\newcommand{\cV}{{\ensuremath{\mathcal{V}}}}

\newcommand{\vd}{{\ensuremath{{\mathbf{d}}}}}

\newcommand{\vl}{{\ensuremath{{\mathbf{l}}}}}

\newcommand{\vp}{{\ensuremath{{\mathbf{p}}}}}

\newcommand{\vr}{{\ensuremath{{\mathbf{r}}}}}

\newcommand{\mC}{{\ensuremath{\mathbf{C}}}}

\newcommand{\mD}{{\ensuremath{\mathbf{D}}}}

\newcommand{\mI}{{\ensuremath{\mathbf{I}}}}

\newcommand{\mL}{{\ensuremath{\mathbf{L}}}}

\newcommand{\mQ}{{\ensuremath{\mathbf{Q}}}}

\newcommand{\mR}{{\ensuremath{\mathbf{R}}}}

\newcommand{\mS}{{\ensuremath{\mathbf{S}}}}

\newcommand{\mU}{{\ensuremath{\mathbf{U}}}}

\newcommand{\mW}{{\ensuremath{\mathbf{W}}}}

\newcommand{\mX}{{\ensuremath{\mathbf{X}}}}

\usepackage{latexsym}

\def\IC{\mathbb C}
\def\IN{\mathbb N}
\def\IZ{\mathbb Z}
\def\IR{\mathbb R}

\def\shat{^{\mathchoice{}{}%
 {\,\,\smash{\hbox{\lower4pt\hbox{$\widehat{\null}$}}}}%
 {\,\smash{\hbox{\lower3pt\hbox{$\hat{\null}$}}}}}}

\def\bSigma{{
      \ooalign{
      \smash{\hskip.4pt\raise.4pt\hbox{$\Sigma$}}\vphantom{}\crcr
      \smash{\hskip.7pt\raise.6pt\hbox{$\Sigma$}}\vphantom{}\crcr
      \smash{\hbox{$\Sigma$}}\vphantom{$\Sigma$}}
      \vphantom{\hbox{$\Sigma$}}
      }}
\def\bTheta{{
      \ooalign{
      \smash{\hskip.5pt\raise.5pt\hbox{$\Theta$}}\vphantom{}\crcr
      \smash{\hskip.0pt\raise.1pt\hbox{$\Theta$}}\vphantom{}\crcr
      \smash{\hbox{$\Theta$}}\vphantom{$\Theta$}}
      \vphantom{\hbox{$\Theta$}}
      }}
\def\bDelta{{
      \ooalign{
      \smash{\hskip.4pt\raise.4pt\hbox{$\Delta$}}\vphantom{}\crcr
      \smash{\hskip.7pt\raise.6pt\hbox{$\Delta$}}\vphantom{}\crcr
      \smash{\hbox{$\Delta$}}\vphantom{$\Delta$}}
      \vphantom{\hbox{$\Delta$}}
      }}
\def\bLambda{{
      \ooalign{
      \smash{\hskip.5pt\raise.5pt\hbox{$\Lambda$}}\vphantom{}\crcr
      \smash{\hskip.0pt\raise.1pt\hbox{$\Lambda$}}\vphantom{}\crcr
      \smash{\hbox{$\Lambda$}}\vphantom{$\Lambda$}}
      \vphantom{\hbox{$\Lambda$}}
      }}

\makeatletter

\def\bordermatrix#1{\begingroup \m@th
  \@tempdima 8.75\p@
  \setbox\z@\vbox{%
    \def\cr{\crcr\noalign{\kern2\p@\global\let\cr\endline}}%
    \ialign{$##$\hfil\kern2\p@\kern\@tempdima&\thinspace\hfil$##$\hfil
      &&\quad\hfil$##$\hfil\crcr
      \omit\strut\hfil\crcr\noalign{\kern-\baselineskip}%
      #1\crcr\omit\strut\cr}}%
  \setbox\tw@\vbox{\unvcopy\z@\global\setbox\@ne\lastbox}%
  \setbox\tw@\hbox{\unhbox\@ne\unskip\global\setbox\@ne\lastbox}%
  \setbox\tw@\hbox{$\kern\wd\@ne\kern-\@tempdima\left[\kern-\wd\@ne
    \global\setbox\@ne\vbox{\box\@ne\kern2\p@}%
    \vcenter{\kern-\ht\@ne\unvbox\z@\kern-\baselineskip}\,\right]$}%
  \null\;\vbox{\kern\ht\@ne\box\tw@}\endgroup}
\makeatother

\makeatletter
\def\argmin{\mathop{\operator@font arg\,min}}
\def\argmax{\mathop{\operator@font arg\,max}}
\makeatother

\newcommand{\diag}{\mbox{\rm diag}}

\newcommand{\tr}{\mbox{\rm tr}}

\def\bm#1{\mbox{\boldmath $#1$}}

\newcommand{\blambda}{{\bm\lambda}}

\newcommand{\bea}{\begin{array}}
\newcommand{\ena}{\end{array}}
\newcommand{\beq}{\begin{equation}}
\newcommand{\enq}{\end{equation}}

\newcommand{\beqa}{\begin{eqnarray}}
\newcommand{\enqa}{\end{eqnarray}}

\newcommand{\beqan}{\begin{eqnarray*}}
\newcommand{\enqan}{\end{eqnarray*}}

\newcommand{\AL}{\begin{enumerate}}
\newcommand{\ALE}{\end{enumerate}}

\def\addots{\mathinner{
    \mkern1mu\raise0pt\vbox{\kern7pt\hbox{.}}
    \mkern2mu\raise4pt\hbox{.}
    \mkern2mu\raise7pt\hbox{.}
    \mkern1mu}}

\def\sddots{\mathinner{
    \mkern.8mu\raise7pt\hbox{.}
    \mkern.8mu\raise4pt\hbox{.}
    \mkern.8mu\raise0pt\vbox{\kern7pt\hbox{.}}
    \mkern1mu}}

\def\saddots{\mathinner{
    \mkern.2mu\raise0pt\vbox{\kern7pt\hbox{.}}
    \mkern.2mu\raise4pt\hbox{.}
    \mkern.2mu\raise7pt\hbox{.}
    \mkern1mu}}

\def\sqplus{\mathbin{
	{\ooalign{\hfil\raise.3ex\hbox{\scriptsize
	+}\hfil\crcr\mathhexbox274\crcr\mathhexbox275}}
	}} 
\def\sqminus{\mathbin{
	{\ooalign{\hfil\raise.3ex\hbox{\scriptsize
	--}\hfil\crcr\mathhexbox274\crcr\mathhexbox275}}
	}}

\def\IC{{
   \mathord{
      \hbox to 0em{
	 \hskip-4pt
         \ooalign{
	   \smash{\hskip1.9pt\raise2.6pt\hbox{$\scriptscriptstyle |$}}\crcr
	   \smash{\hbox{\rm\sf C}} }
	 \hidewidth}
      \phantom{\hbox{\rm\sf C}}
} }}
\def\IN{
    {\ooalign{
   \smash{\hskip2.2pt\raise1.5pt\hbox{$\scriptscriptstyle |$}}\vphantom{}\crcr
   \hbox{\sf N}
	}}
	} 
\def\IZ{
    {\ooalign{
   \smash{\hskip1.9pt\raise0pt\hbox{$\sf Z$}}\vphantom{}\crcr
   \hbox{\sf Z}
	}}
	} 
\def\IR{
    {\ooalign{
   \smash{\hskip2.2pt\raise1.5pt\hbox{$\scriptscriptstyle |$}}\vphantom{}\crcr
   \smash{\hskip2.2pt\raise3.3pt\hbox{$\scriptscriptstyle |$}}\vphantom{}\crcr
   \hbox{\sf R}
	}}
	} 

\DeclareMathAlphabet{\mathcmb}{OT1}{cmr}{b}{n}

\def\bSigma{\ensuremath{\mathcmb{\Sigma}}}
\def\bLambda{\ensuremath{\mathcmb{\Lambda}}}

\def\bTheta{\ensuremath{\mathcmb{\Theta}}}

\newcommand{\SI}{\begin{indlist}}
\newcommand{\EI}{\end{indlist}}

\newcommand{\DL}{\begin{dashlist}}
\newcommand{\DLE}{\end{dashlist}}

\makeatletter
\def\setboxz@h{\setbox\z@\hbox}
\def\wdz@{\wd\z@}
\def\boxz@{\box\z@}
\def\underset#1#2{\binrel@{#2}%
  \binrel@@{\mathop{\kern\z@#2}\limits_{#1}}}
\def\binrel@#1{\begingroup
  \setboxz@h{\thinmuskip0mu
    \medmuskip\m@ne mu\thickmuskip\@ne mu
    \setbox\tw@\hbox{$#1\m@th$}\kern-\wd\tw@
    ${}#1{}\m@th$}%
  \edef\@tempa{\endgroup\let\noexpand\binrel@@
    \ifdim\wdz@<\z@ \mathbin
    \else\ifdim\wdz@>\z@ \mathrel
    \else \relax\fi\fi}%
  \@tempa
}
\let\binrel@@\relax%
\makeatother